\documentclass[11pt]{article}
\usepackage[utf8]{inputenc}
\usepackage[T1]{fontenc}
\usepackage{lmodern}
\usepackage{amsmath,amssymb}
\usepackage{graphicx}
\usepackage{booktabs}
\usepackage{geometry}
\usepackage{siunitx}
\usepackage[numbers,sort&compress]{natbib}
\usepackage{url}  %  \urlstyle/\Url .bbl  \doi  catcode DOI 
\usepackage{placeins}  % \FloatBarrier/
\usepackage{caption}
\usepackage{xcolor}
\newcommand{\Rwear}{56\%}        %   total-wear reduction
\newcommand{\Rpeak}{59\%}        %  peak-wear reduction
\newcommand{\Rflat}{28\%}        %  flatness improvement
\newcommand{\Rvrip}{58\%}        %  velocity-ripple improvement
\newcommand{\Rstep}{19\%}        %  step-length increase
\newcommand{\Rclear}{15\%}       %  clearance reduction
\newcommand{\Rlink}{29\%}        %  max link-length change
\newcommand{\Rbandlo}{48\%}      %  
\newcommand{\Rbandhi}{56\%}      %  
\newcommand{\Rcrankshare}{14\%}  % 
\newcommand{\Rmferrlo}{3\%}      %   (Jansen)
\newcommand{\Rmferrhi}{4\%}      %   ()
\newcommand{\Rwearint}{56\%}     % 
\newcommand{\Rlife}{2.3}         % (1/)
\newcommand{\flatJ}{0.0281}\newcommand{\flatO}{0.0204}
\newcommand{\vripJ}{0.0956}\newcommand{\vripO}{0.0406}
\newcommand{\stepJ}{43.3}\newcommand{\stepO}{51.6}
\newcommand{\clearJ}{25.7}\newcommand{\clearO}{21.9}
\newcommand{\dutyJ}{0.202}\newcommand{\dutyO}{0.186}
\newcommand{\Rduty}{8\%}          % 
\newcommand{\Rfonelo}{0.56}\newcommand{\Rfonehi}{0.57}  % f1 
\newcommand{\Rftwolo}{0.44}\newcommand{\Rftwohi}{0.46}  % f2 
\newcommand{\Rfrontwearlo}{54}    %  (%) \Rwear 
\newcommand{\Rfloornine}{54\%}      %  0.90Jansen  
\newcommand{\Rfloorninefive}{49\%}  %  0.95Jansen  
\newcommand{\HVa}{0.357}\newcommand{\HVb}{0.279}\newcommand{\HVc}{0.275}  % (3,densify)
\newcommand{\HVstd}{0.038}        % HV
\newcommand{\Rhvgain}{18\%}       % NSGA-II   
\newcommand{\Nrandsamp}{2500}     %  
\newcommand{\Nrandfeas}{243}      %  
\newcommand{\Rverifytorque}{0.01\%}  %   
\newcommand{\Rverifypower}{3\%}      %  
\newcommand{\RsobolcST}{0.85}\newcommand{\RsobolcSone}{0.15}  % c /
\newcommand{\RsobolkST}{0.76}\newcommand{\RsobolkSone}{0.39}  % k /
\newcommand{\Rtolfeas}{49\%}      % 1% 
\newcommand{\Rtolmean}{43\%}      % 1% 
\newcommand{\Rtolpfive}{24\%}        % 1% 5
\newcommand{\Rtoltwo}{32\%}       % 2% 

\title{\textbf{Durability-Aware Multi-Objective Optimization of the Jansen
Linkage: Trading Gait Quality Against Joint Wear}}
\author{Jichao Wang\\ James Watt School of Engineering,
University of Glasgow\\ Glasgow, United Kingdom\\
\texttt{jichaowang02@gmail.com}\\
ORCID: 0009-0000-6551-6006}
\date{\today}

\begin{document}
\maketitle

\begin{abstract}
\noindent\textbf{Abstract.}
The Jansen linkage is a single-degree-of-freedom planar leg mechanism whose
eleven ``holy numbers'' were evolved by Theo Jansen to optimize the foot-path
\emph{gait} alone, with no regard for the wear of its revolute joints. This
paper introduces a durability objective into the design of the Jansen leg. A
parametric forward-kinematic model (two-circle-intersection solver), an
inverse-dynamic model (constraint-Jacobian / Lagrange-multiplier formulation of
a seven-body, ten-joint system, independently cross-verified by a reduced-DOF
energy method), and an Archard wear model are coupled to
evaluate, for any set of link lengths, both gait quality and the per-cycle
sliding wear at every pin. Because the wear is computed on \emph{ideal,
clearance-free} revolute joints, the resulting wear figures are a
\emph{relative} comparative ranking rather than an absolute life prediction.
A bi-objective problem---composite gait error versus total joint wear, subject
to step-length, ground-clearance, duty-factor and assembly constraints---is
solved with NSGA-II. Under the adopted gait metric the classical Jansen design
is \emph{Pareto-dominated}: for a representative design, link-length
adjustments within $\pm\Rlink$ simultaneously flatten the stance ($-\Rflat$),
smooth the stance velocity ($-\Rvrip$) and reduce total joint wear by
$\sim\Rwear$. A sensitivity study shows the wear advantage is robust across a
crank-speed$\,\times\,$payload envelope ($\Rbandlo$--$\Rbandhi$) and identifies
the link lengths that most strongly govern wear. A variance-based global (Sobol)
analysis confirms that two link lengths dominate the wear variance, and a
Monte-Carlo manufacturing-tolerance study shows the wear advantage degrades
gracefully under realistic fabrication error. The framework provides a
practical route to longer-lived walking linkages and a baseline for future
wear--clearance--impact coupled studies.

\par\medskip
\noindent\textbf{Keywords:} Jansen linkage; walking mechanism;
multibody dynamics; Archard wear; multi-objective optimization; NSGA-II
\end{abstract}

\section*{Nomenclature}
\noindent
\begin{tabular}{ll}
$a$--$m$ & link lengths (holy numbers), mm \\
$\theta,\omega$ & crank angle and angular speed \\
$O,G$ & crank-centre and frame ground pivots \\
$J_1$--$J_5,F$ & moving joints and the foot point \\
$q,\dot q,\ddot q$ & generalized coordinates \\
$\mathbf{M},\Phi_q,\lambda$ & mass matrix, constraint Jacobian, multipliers \\
$\mathbf{Q},W$ & generalized applied force; stance ground load \\
$k,\bar F,s,r_{\mathrm{pin}}$ & wear coeff., mean pin force, sliding, pin radius \\
$\Delta\phi$ & per-cycle relative rotation at a pin \\
$f_1,f_2$ & gait-error and total-wear objectives \\
\end{tabular}

\section{Introduction}

Walking linkages convert a single rotary input into a foot trajectory with a
near-straight ground-contact stroke and a high return arc, enabling legged
locomotion without per-joint actuation. The Jansen linkage, popularized through
Theo Jansen's \emph{Strandbeest}, is the best-known example; its eleven link
lengths---the ``holy numbers''---were obtained by an evolutionary search scored
only on gait criteria such as foot-path flatness, ground clearance and the
absence of cusps~\citep{jansen}. Research on the mechanism has focused almost
exclusively on kinematics, motor torque and gait: Patnaik and
Umanand~\citep{patnaik2015} derived its kinematics and dynamics via a bond-graph
model and the circle-intersection method, and Nansai et
al.~\citep{nansai2015} studied actuation and control of a Jansen-based quadruped.
Dimensional optimization of the leg has likewise targeted the gait trajectory
alone~\citep{zang2017} (with variants repurposed for gait
rehabilitation~\citep{mohanvarma2021}), and other single-DOF leg linkages---the Klann
mechanism~\citep{kavlak2021} and novel eight-link designs~\citep{desai2019}---have
been analyzed on the same kinematic footing.

In contrast, the durability of the revolute joints---which ultimately limits
the service life of any physical walking machine---has not been considered in the
design of the link lengths. Joint wear is governed by the Archard
law~\citep{archard1953,archard1956}, the product of contact load and relative
sliding distance. In mechanisms it is commonly predicted by integrating a
multibody-dynamics model with a wear model, frequently including clearance-joint
contact~\citep{flores2004,tian2018}: Flores~\citep{flores2009} and Mukras et
al.~\citep{mukras2010} established revolute-clearance-joint wear prediction, since
validated experimentally and cast as integrated dynamics--wear
loops~\citep{bai2014,lai2017}; recent work~\citep{jia2025,liu2025,chenwang2025} further couples
wear with rigid--flexible and tribo-dynamic clearance models. Multi-objective
evolutionary search is by now standard for dimensional synthesis, trading
path-tracking error against transmission-angle or velocity
criteria~\citep{narimanzadeh2009,khorshidi2011,sleesongsom2018,kong2010,lee2024}, and
genetic algorithms have optimized link parameters under joint clearance to reduce
path error~\citep{erkaya2009}---but joint wear itself has rarely been adopted as
an explicit design objective, and to our knowledge never for the Jansen leg. Moreover, these studies
treat generic slider-crank, four-bar or multi-link mechanisms driven at constant
speed---never the gait-specialized, intermittently ground-loaded Jansen leg.

This paper closes that gap. We build a fully parametric pipeline that, for
any set of link lengths, evaluates (i) gait quality and (ii) the per-cycle wear
at every pin joint, by coupling forward kinematics, inverse dynamics and the
Archard law, and pose a bi-objective optimization---gait quality versus total
joint wear---solved with NSGA-II. The contributions are: (1) a parametric
kinematic--dynamic--wear evaluator for the Jansen leg, cross-verified by an
independent reduced single-DOF energy formulation; (2) the finding that
Jansen's holy numbers are Pareto-dominated once durability is considered, with
optimized designs that improve gait \emph{and} reduce wear; (3) a robustness and
sensitivity analysis identifying the wear-governing link lengths and confirming
the conclusion across operating conditions.

\section{Kinematic Model}

\begin{figure}[htbp]
\centering
\includegraphics[width=0.6\linewidth]{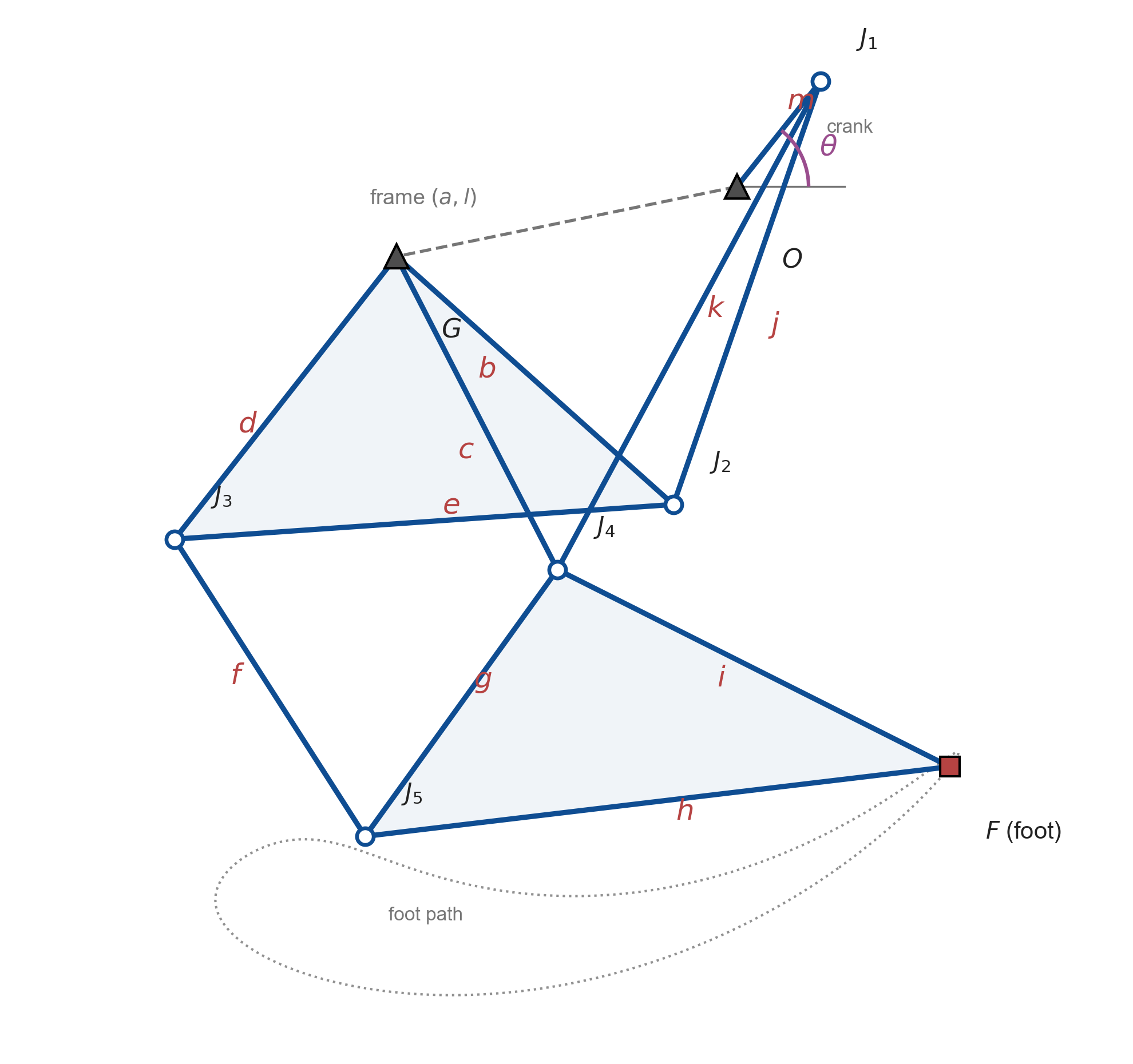}
\caption{Schematic of the Jansen leg: linkage topology and notation---link
lengths $a$--$m$ (the ``holy numbers''), revolute joints $J_1$--$J_5$, foot point
$F$, fixed frame pivots $O,G$, and the crank input angle $\theta$.}
\label{fig:schematic}
\end{figure}

The Jansen leg comprises a crank and a linkage forming seven moving rigid
bodies connected by ten revolute joints (the topology and notation are shown in
Fig.~\ref{fig:schematic}). Including the ground link ($n=8$,
$j=10$ joints), the Gr\"ubler count gives $3(n-1)-2j = 3\times 7 - 20 = 1$
degree of freedom. Two ground pivots are fixed: the crank centre $O=(0,0)$ and
the frame pivot $G=(-a,-l)$. With crank angle $\theta$ as input, the crank tip
is $J_1=O+m(\cos\theta,\sin\theta)$, and the remaining joints follow as
two-circle intersections (link lengths in Table~\ref{tab:holy}):
\begin{align}
J_2 &= \mathrm{circ}(J_1,j;\,G,b), &
J_3 &= \mathrm{circ}(J_2,e;\,G,d), \\
J_4 &= \mathrm{circ}(J_1,k;\,G,c), &
J_5 &= \mathrm{circ}(J_3,f;\,J_4,g), \\
F   &= \mathrm{circ}(J_4,i;\,J_5,h),
\end{align}
where $\mathrm{circ}(P_1,r_1;P_2,r_2)$ is the branch-selected intersection of
circles of radii $r_1,r_2$ centred at $P_1,P_2$, and $F$ (the foot) is the apex
of the rigid triangle $J_4 J_5 F$.

\begin{table}[htbp]
\centering
\caption{Jansen ``holy numbers'' (link lengths, mm)}
\label{tab:holy}
\small
\begin{tabular}{lccccccccccccc}
\toprule
 & $a$ & $b$ & $c$ & $d$ & $e$ & $f$ & $g$ & $h$ & $i$ & $j$ & $k$ & $l$ & $m$ \\
\midrule
Value & 38.0 & 41.5 & 39.3 & 40.1 & 55.8 & 39.4 & 36.7 & 65.7 & 49.0 & 50.0 & 61.9 & 7.8 & 15.0 \\
\bottomrule
\end{tabular}
\end{table}

\begin{figure}[htbp]
\centering
\includegraphics[width=0.95\linewidth]{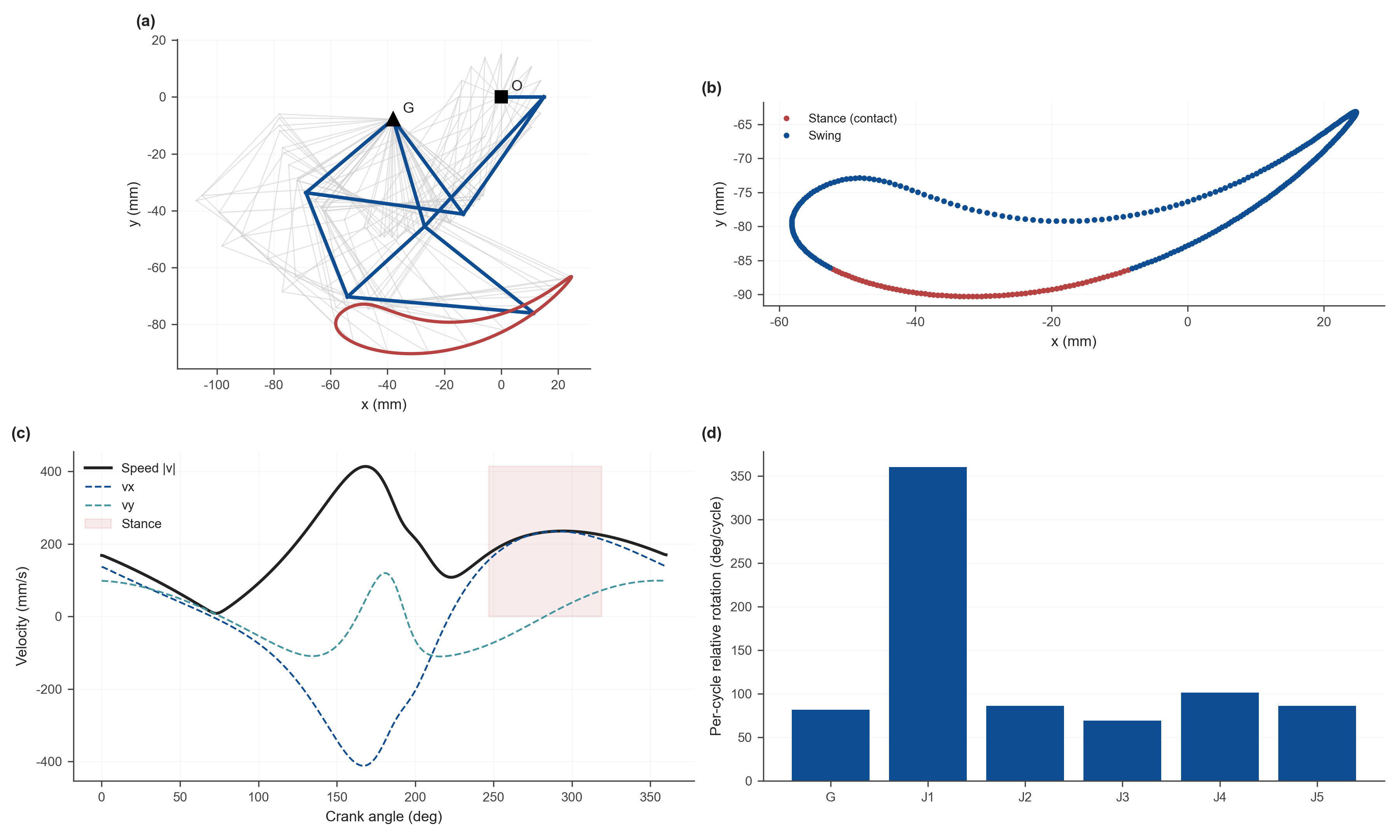}
\caption{Kinematic analysis: linkage poses and foot trajectory; stance/swing
phases; foot velocity; per-pin relative rotation.}
\label{fig:kin}
\end{figure}

Sweeping $\theta\in[0,2\pi)$ yields the foot trajectory
(Fig.~\ref{fig:kin}), which reproduces the characteristic Jansen gait---a long,
nearly flat ground-contact stroke followed by a high return arc---with a stance
duty factor (foot in the lowest $15\%$ of its height range) of $\approx 20\%$,
validating the kinematic model qualitatively.

\section{Dynamic Model}

Since the kinematics fully determines the configuration $q$, velocity
$\dot q$ and acceleration $\ddot q$ of all seven bodies (periodic finite
differences in time $t=\theta/\omega$), the joint reaction forces follow by
\emph{inverse} dynamics. With $q=[x_i,y_i,\varphi_i]_{i=1}^{7}\in\mathbb{R}^{21}$
and the holonomic revolute constraints $\Phi(q,t)=0$~\citep{shabana2020} (two per joint plus the
driving constraint $\varphi_{\mathrm{crank}}=\omega t$, $21$ equations), the
Newton--Euler equations are
\begin{equation}
\mathbf{M}\ddot q + \Phi_q^{\mathsf T}\lambda = \mathbf{Q},
\qquad
\lambda = \big(\Phi_q^{\mathsf T}\big)^{-1}\big(\mathbf{Q}-\mathbf{M}\ddot q\big),
\label{eq:invdyn}
\end{equation}
with $\mathbf{M}=\mathrm{diag}(m_i,m_i,I_i)$ the mass matrix (uniform-slender-bar
masses/inertias from the line density in Table~\ref{tab:param}), $\Phi_q$ the
$21\times 21$ constraint Jacobian, $\mathbf{Q}$ the applied generalized forces
(gravity and, during stance, the vertical ground-reaction load $W$ at the foot),
and $\lambda$ the Lagrange multipliers; each multiplier pair is the reaction
force at the corresponding pin and the driving-constraint multiplier is the
crank torque. The revolute joints are thus treated as ideal, clearance-free
holonomic constraints---no pin--bore contact-force law is invoked---and the
stance load is a prescribed vertical force rather than a continuous foot--ground
contact model; both idealizations are revisited as limitations in
Section~7. Should the leg pass near a singular configuration where
$\Phi_q^{\mathsf T}$ loses rank, that instant is handled by a least-squares
solution of \eqref{eq:invdyn} and excluded from the cycle means. For both
designs reported here this safeguard is never triggered: all $361$ crank angles
are well-conditioned ($\mathrm{cond}\,\Phi_q^{\mathsf T}<10^3$), so no angle is
excluded and the cycle-mean reaction forces---and hence the wear figures---are
unaffected by the singular-angle handling.

\begin{table}[htbp]
\centering
\caption{Baseline model parameters}
\label{tab:param}
\small
\begin{tabular}{lll}
\toprule
Parameter & Value & Note \\
\midrule
Line density $\rho_\ell$ & \SI{0.05}{\kilo\gram\per\metre} & uniform slender bars \\
Crank speed $\omega$ & $1$\,rev/s ($2\pi$\,rad/s) & baseline \\
Stance load $W$ & \SI{20}{\newton} & vertical ground reaction \\
Pin radius $r_{\mathrm{pin}}$ & \SI{4}{\milli\metre} & sliding = $\Delta\phi\,r_{\mathrm{pin}}$ \\
Wear coeff. $k$ & \SI{1e-13}{\cubic\metre\per\newton\per\metre} & dimensional, steel-on-steel order \\
\bottomrule
\end{tabular}
\end{table}

\begin{figure}[htbp]
\centering
\includegraphics[width=0.95\linewidth]{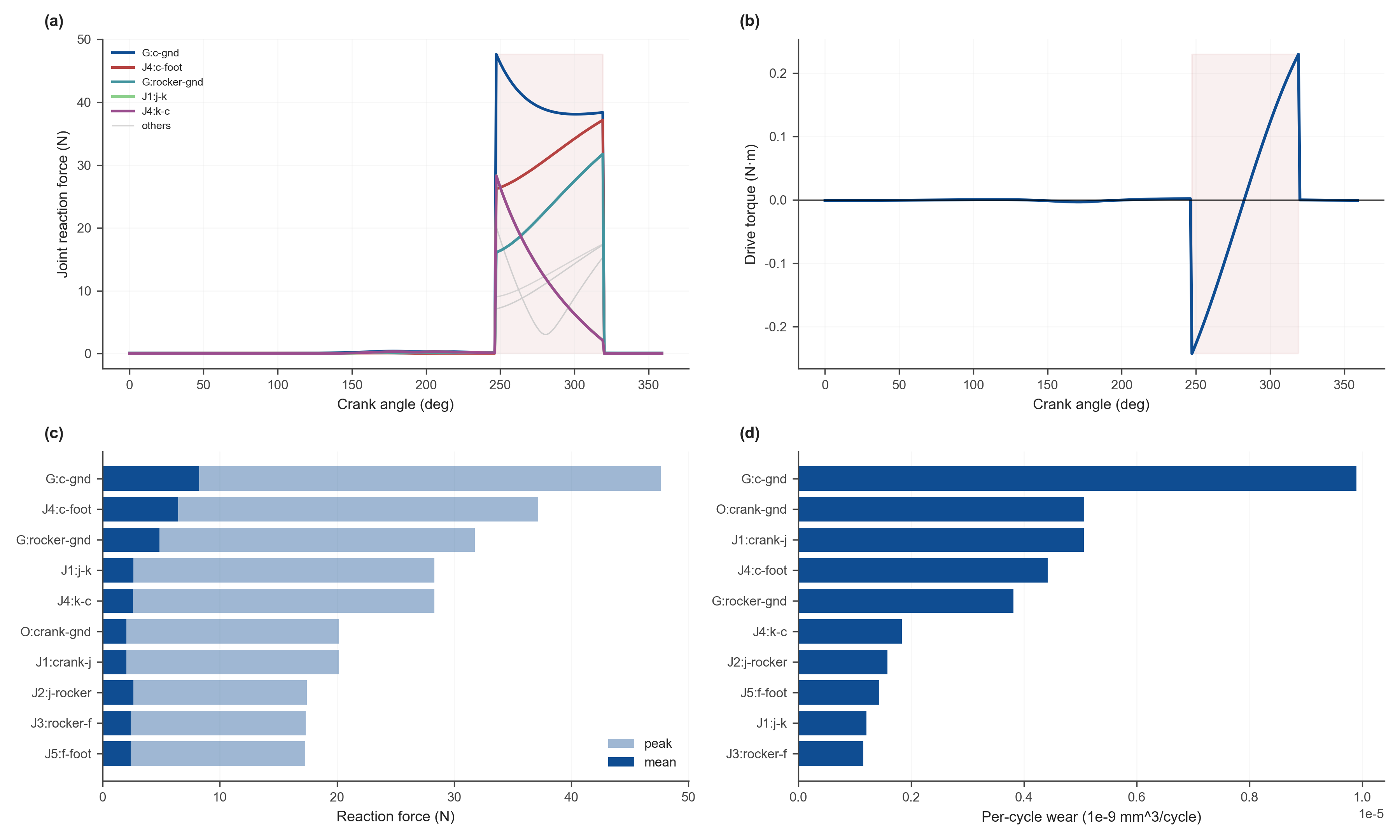}
\caption{Dynamics over the gait cycle: (a) joint reaction forces (stance shaded;
the five highest-peak pins highlighted, the rest in grey); (b) crank input torque;
(c) peak/mean force ranking; (d) Archard wear distribution.}
\label{fig:dyn}
\end{figure}

Solving \eqref{eq:invdyn} over the gait cycle (Fig.~\ref{fig:dyn}) shows
joint reaction forces negligible during swing and rising sharply during stance,
peaking at $17$--$\SI{48}{\newton}$ for $W=\SI{20}{\newton}$; the grounded load-bearing pin
$G$ carries the highest \emph{force} (\SI{47.6}{\newton}). The crank input torque peaks
at $\SI{0.24}{\newton\metre}$ and integrates to $\approx 0$ over a cycle---a
consequence of the purely vertical, conservative load model (gravity returns to
zero; the vertical $W$ does no net work over the closed foot path); a horizontal
traction component would yield net positive propulsive work and a non-zero
cycle-integrated torque.

\section{Wear Model}

Pin-joint wear is modelled by the Archard law: the volume removed at an
interface over one gait cycle is
\begin{equation}
V = k\textstyle\oint F\,\mathrm{d}s \;\approx\; k\,\bar F\,s,
\qquad s = \Delta\phi\,r_{\mathrm{pin}},
\label{eq:archard}
\end{equation}
where $k$ is the \emph{dimensional} wear coefficient ($k=K/H$ with $K$ the
dimensionless Archard coefficient and $H$ the hardness; units
$\si{\cubic\metre\per\newton\per\metre}=\si{\per\pascal}$, here $k=\SI{1e-13}{\cubic\metre\per\newton\per\metre}$, a steel-on-steel
order of magnitude), $\bar F$ the cycle-mean reaction force at the pin
(Section~3), $r_{\mathrm{pin}}$ the pin radius, and $\Delta\phi$ the purely
kinematic per-cycle relative rotation. The mean-force form $\bar F s$ replaces
the exact $\oint F\,\mathrm{d}s$; because force and sliding are phase-correlated
for load-dominated pins (load concentrated in stance, sliding spread over the
cycle), this approximation can bias individual load-dominated pins. We quantified
this: $\bar F s$ agrees with the exact $\oint F\,\mathrm{d}s$ on absolute total
per-cycle wear to within $\Rmferrlo$--$\Rmferrhi$, and the
optimized-to-Jansen \emph{reduction} is essentially identical ($\Rwear$ via $\bar F s$ vs
$\Rwearint$ via $\oint F\,\mathrm{d}s$). The comparative conclusion is therefore
robust to the approximation (and, since $k$ and $r_{\mathrm{pin}}$ cancel in the
ratio, independent of the tribological constants).

Two regimes characteristic of a walking leg emerge. The crank--ground and
crank--coupler pins rotate a full $360^\circ$ each cycle---large sliding $s$ but
modest force, hence \emph{slip-dominated}; crucially their sliding is fixed by the
input rotation and \emph{independent of the link lengths} $b\ldots k$, an
irreducible wear floor that no geometric change can lower. The grounded
load-bearing pins, by contrast, oscillate through small angles but carry the
stance load (\emph{load-dominated}), and the most heavily loaded of them is in
fact the \emph{peak-wear} pin. For the Jansen design the slip-dominated crank
bearing accounts for only $\Rcrankshare$ of the total per-cycle wear, so both the
peak and the bulk of wear reside in the design-sensitive, load-dominated pins.
Total wear---which aggregates this reducible population---is therefore adopted as
the durability objective; peak wear, itself load-dominated, is reported as a
secondary outcome and falls comparably ($\Rpeak$).

\section{Multi-Objective Optimization}

The ten link lengths $b\ldots k$ are the design variables, each bounded to
$\pm 30\%$ of its Jansen value; the frame ($a,l$) and crank ($m$) are fixed so
the frame and rotary input are identical across designs. Two objectives are
minimized,
\begin{equation}
f_1 = \tfrac12\frac{\mathrm{flat}}{\mathrm{flat}_J}
     +\tfrac12\frac{\mathrm{vrip}}{\mathrm{vrip}_J},
\qquad
f_2 = \frac{W_{\mathrm{tot}}}{W_{\mathrm{tot},J}},
\end{equation}
where $f_1$ is a composite gait error (stance flatness plus stance
horizontal-velocity ripple, equally weighted and normalized to the Jansen
baseline $J$) and $f_2$ the normalized total joint wear. Three inequality
constraints require step length, ground clearance and duty factor each
$\ge 0.85$ of the Jansen value; non-assembling designs are infeasible. The
problem is solved with NSGA-II~\citep{deb2002} (pymoo~\citep{blank2020}) using the
settings in Table~\ref{tab:nsga}. Because only $\sim 10\%$ of random designs are
feasible, each run uses a mixed initial population---half sampled uniformly over
the $\pm 30\%$ box (global exploration) and half drawn from a Gaussian
neighbourhood ($\sigma=8\%$) of the Jansen design (local refinement); three
independent seeds are merged into a single non-dominated front.

\begin{table}[htbp]
\centering
\caption{NSGA-II settings}
\label{tab:nsga}
\small
\begin{tabular}{ll}
\toprule
Population & 100 \\
Generations & 80 \\
Independent seeds & 3 (merged \\
Crossover & SBX (pymoo default $\eta,p$) \\
Mutation & polynomial (pymoo default) \\
Initial seeding & mixed: 50\% uniform $+$ 50\% Gaussian ($\sigma=8\%$) \\
Crank-angle samples & 361 per evaluation \\
\bottomrule
\end{tabular}
\end{table}

\section{Results and Discussion}

\begin{figure}[htbp]
\centering
\includegraphics[width=0.62\linewidth]{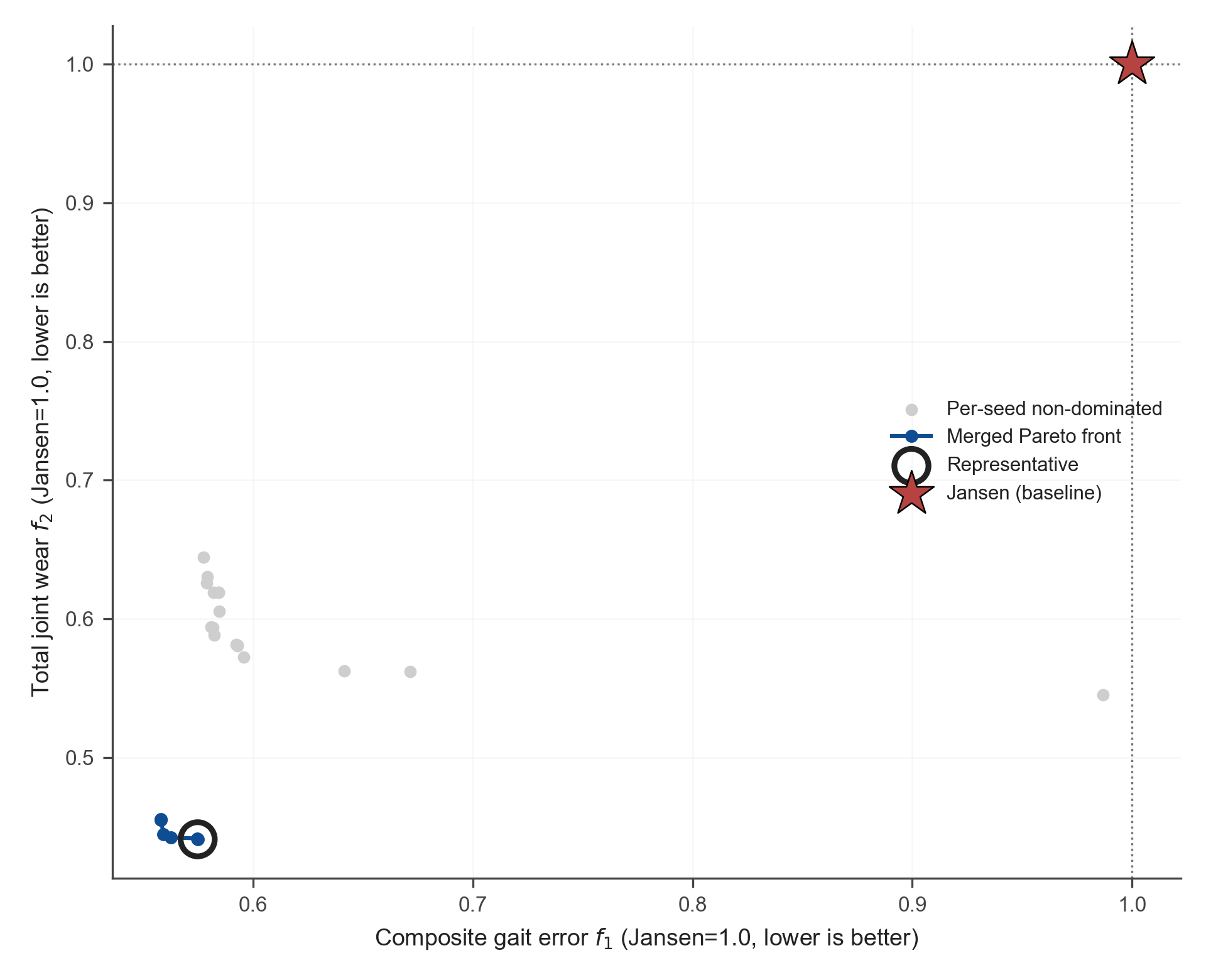}
\caption{Merged Pareto front (blue) of total joint wear vs composite gait error,
both normalized to Jansen $=1$; per-seed non-dominated points in grey, the chosen
representative ringed. Jansen (red star) is dominated by the entire front.}
\label{fig:pareto}
\end{figure}

\textbf{Pareto front.} Fig.~\ref{fig:pareto} shows the merged front (seven
non-dominated designs from three seeds) in the gait--wear plane, Jansen
normalized to $(1,1)$. The entire front lies below and to the left of the Jansen
point ($f_1\in[\Rfonelo,\Rfonehi]$, $f_2\in[\Rftwolo,\Rftwohi]$): the classical design is
dominated. This dominance is \emph{weight-independent}---all seven front designs
improve flatness \emph{and} stance-velocity ripple \emph{and} total wear
simultaneously, so they dominate Jansen for any flatness/ripple weighting (we
verified $f_1$ weights of $0.3$, $0.5$, $0.7$ all leave Jansen dominated by all
seven). Along the front, total wear is reduced by $\Rfrontwearlo$--$\Rwear$ while gait
also improves; the narrow extent of the front indicates that gait quality and
durability are largely \emph{compatible} rather than strongly conflicting in
this design space. All non-dominated designs require a comparable link-length
change (up to $\le\Rlink$); smaller perturbations of Jansen remain dominated and
off the front.

\begin{figure}[htbp]
\centering
\includegraphics[width=0.95\linewidth]{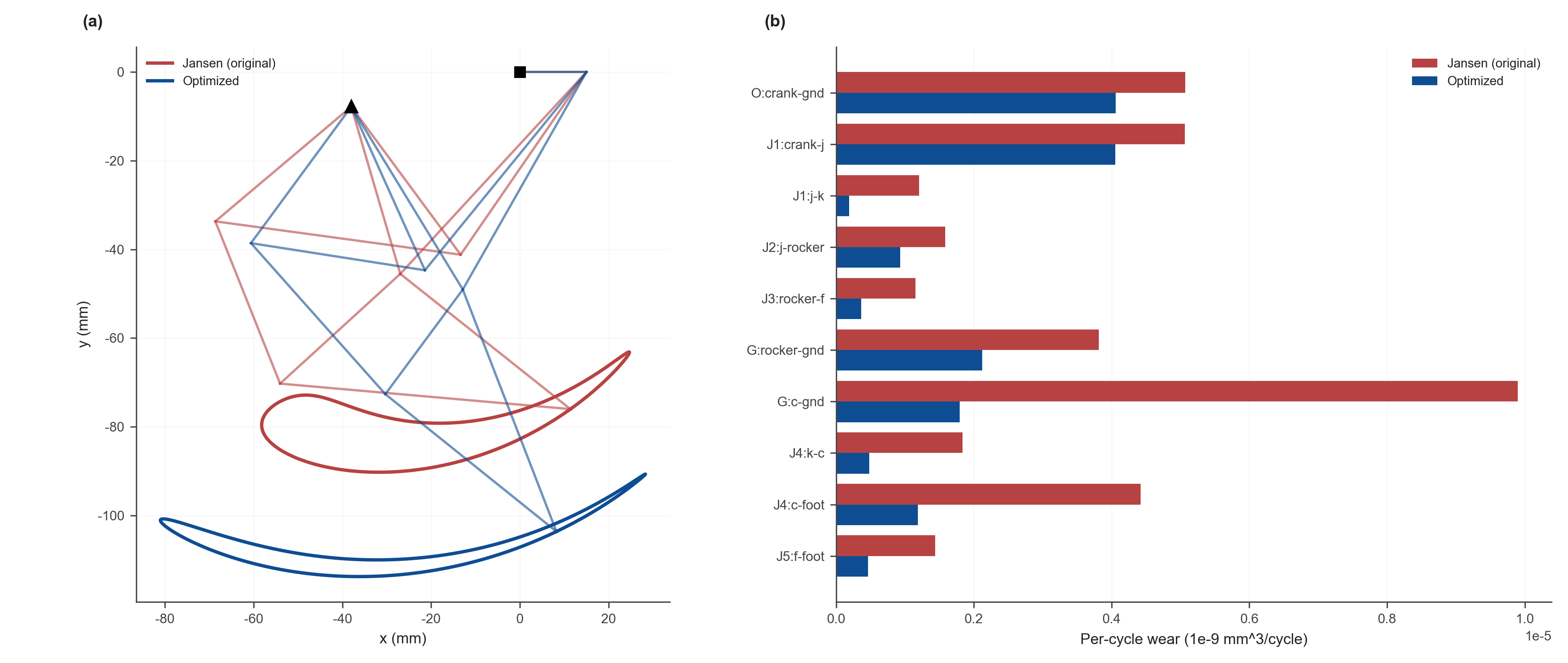}
\caption{Optimized design vs.\ Jansen: foot path and linkage (left); per-joint
wear (right). The optimized leg is a modest ($\le\Rlink$) redesign that lowers
wear at every pin.}
\label{fig:cmp}
\end{figure}

\textbf{Optimized vs.\ Jansen.} Table~\ref{tab:cmp} and Fig.~\ref{fig:cmp}
compare a representative design (selected as the minimum-wear point with
gait error no worse than Jansen) against Jansen; its optimized link lengths are
given in Table~\ref{tab:lengths}. With every link length changed by at most
$\Rlink$---so the optimized leg remains a sensible Jansen-type mechanism, not a
metric exploit---stance flatness improves by $\Rflat$, stance velocity ripple by
$\Rvrip$ and step length by $\Rstep$, while total joint wear falls $\Rwear$ and
peak wear $\Rpeak$. The cost is a $\Rclear$ reduction in ground clearance; this
constraint is \emph{active} (the representative design sits near the $0.85$
clearance floor). Re-optimizing under stricter clearance floors shows the wear
benefit is robust to this design choice rather than an artifact of it: with the
floor raised to $0.90$ and $0.95\times$ the Jansen clearance, total wear still
falls by $\Rfloornine$ and $\Rfloorninefive$ respectively (versus $\Rwear$ at
$0.85$), and Jansen remains Pareto-dominated throughout. The per-joint comparison (Fig.~\ref{fig:cmp}, right) shows wear
decreasing at \emph{every} pin ($-20\%$ to $-85\%$; Table~\ref{tab:perjoint}), a
broad-based improvement rather than a single-joint artifact. For a fixed allowable wear depth on the idealized
clearance-free joints, the $\Rwear$ cut in per-cycle wear projects to roughly a
$\Rlife\times$ gain in joint service life---an idealized relative upper bound
under unchanged material and contact assumptions, not an absolute life
prediction.

\begin{table}[htbp]
\centering
\caption{Optimized design vs.\ Jansen (fair multi-criteria comparison)}
\label{tab:cmp}
\small
\begin{tabular}{lccc}
\toprule
Metric & Jansen & Optimized & Change \\
\midrule
Stance flatness & \flatJ & \flatO & $-\Rflat$ \\
Velocity ripple & \vripJ & \vripO & $-\Rvrip$ \\
Step length & \stepJ & \stepO & $+\Rstep$ \\
Ground clearance & \clearJ & \clearO & $-\Rclear$ \\
Duty factor & \dutyJ & \dutyO & $-\Rduty$ \\
\textbf{Total wear }    & \multicolumn{2}{c}{---} & \textbf{$-$\Rwear} \\
Peak wear & \multicolumn{2}{c}{---} & $-\Rpeak$ \\
\bottomrule
\end{tabular}
\end{table}

\begin{table}[htbp]
\centering
\caption{Optimized link lengths of the representative design (mm)}
\label{tab:lengths}
\small
\begin{tabular}{lcccccccccc}
\toprule
 & $b$ & $c$ & $d$ & $e$ & $f$ & $g$ & $h$ & $i$ & $j$ & $k$ \\
\midrule
Jansen    & 41.5 & 39.3 & 40.1 & 55.8 & 39.4 & 36.7 & 65.7 & 49.0 & 50.0 & 61.9 \\
Optimized & 40.5 & 48.3 & 38.2 & 39.7 & 45.6 & 29.4 & 49.5 & 58.5 & 57.6 & 56.4 \\
Change (\%) & $-2$ & $+23$ & $-5$ & $-29$ & $+16$ & $-20$ & $-25$ & $+19$ & $+15$ & $-9$ \\
\bottomrule
\end{tabular}
\end{table}

\textbf{A family of front designs.} The representative is one point on a
tightly clustered Pareto front; Table~\ref{tab:family} lists four designs spanning
it, from a gait-leaning end (D1) to the wear-leaning representative (D4).
\emph{Every} front design improves all gait metrics \emph{and} cuts total wear (by
$\Rfrontwearlo$--$\Rwear$), within $\le\Rlink$ link-length change. The front is
narrow---across it, wear reduction varies by only $\sim2$ points
($\Rfrontwearlo$--$\Rwear$) while velocity-ripple reduction stays at $58$--$60\%$---so
gait quality and durability are not in tension in this design space: a single
redesign secures both.

\begin{table}[htbp]
\centering
\caption{A family of Pareto-front designs (changes vs.\ Jansen)}
\label{tab:family}
\small
\begin{tabular}{lcccccccc}
\toprule
Design & $f_1$ & $f_2$ & Wear & Flatness & Ripple & Step & Clear. & Max $\Delta L$ \\
\midrule
D1 (gait-leaning) & 0.558 & 0.456 & $-54\%$ & $-29\%$ & $-60\%$ & $+17\%$ & $-15\%$ & $\le29\%$ \\
D2 & 0.558 & 0.455 & $-55\%$ & $-29\%$ & $-60\%$ & $+17\%$ & $-15\%$ & $\le29\%$ \\
D3 & 0.563 & 0.442 & $-56\%$ & $-29\%$ & $-59\%$ & $+18\%$ & $-15\%$ & $\le29\%$ \\
D4 (wear-leaning) & 0.575 & 0.441 & $-\Rwear$ & $-\Rflat$ & $-\Rvrip$ & $+\Rstep$ & $-\Rclear$ & $\le\Rlink$ \\
\bottomrule
\end{tabular}
\end{table}

\begin{figure}[htbp]
\centering
\includegraphics[width=0.95\linewidth]{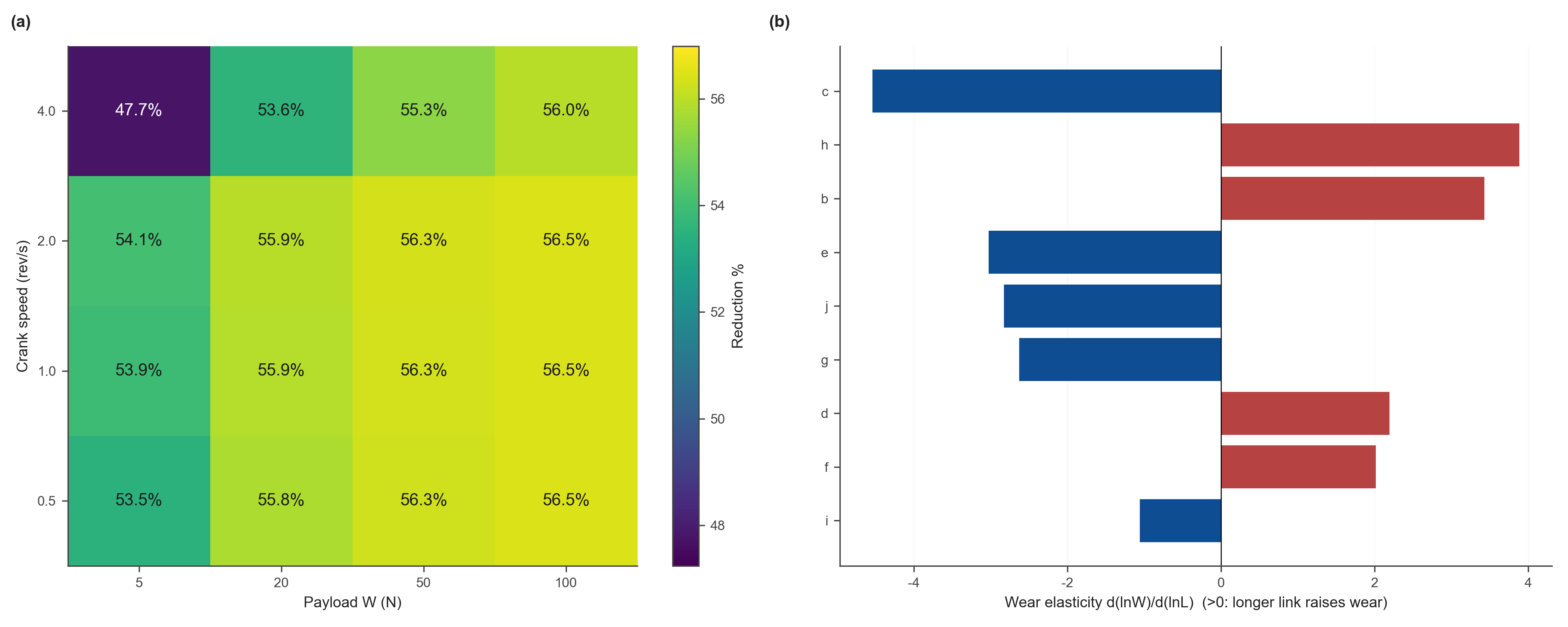}
\caption{Sensitivity: (left) total-wear reduction across a crank-speed$\times$
payload envelope; (right) elasticity of total wear to each link length.}
\label{fig:sens}
\end{figure}

\begin{figure}[htbp]
\centering
\includegraphics[width=0.7\linewidth]{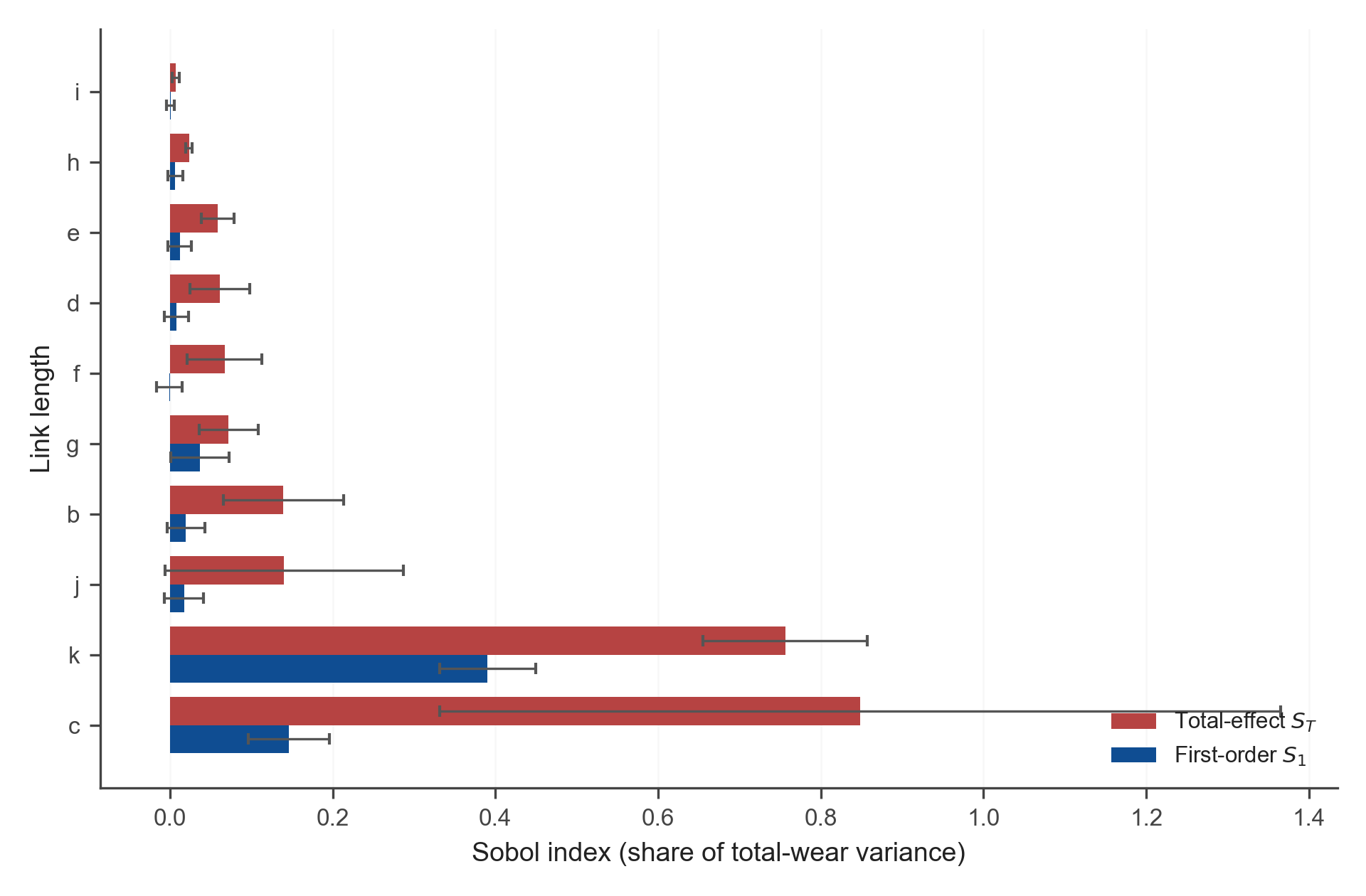}
\caption{Variance-based global sensitivity (Sobol) of normalized total joint wear
to each link length, over a $\pm10\%$ box about the Jansen design: first-order
($S_1$) and total-effect ($S_T$) indices ($N=2048$; bars show 95\% confidence
intervals).}
\label{fig:sobol}
\end{figure}

\textbf{Robustness and sensitivity.} Fig.~\ref{fig:sens} (left) sweeps crank
speed ($0.5$--$4$\,rev/s) against payload ($5$--$\SI{100}{\newton}$): the wear
reduction stays in the $\Rbandlo$--$\Rbandhi$ band throughout, dipping only at
high speed and low load where inertia dominates. Because $k$ and $r_{\mathrm{pin}}$
cancel in the optimized-to-Jansen ratio, the conclusion is independent of the
(uncertain) tribological constants. A local elasticity study
(Fig.~\ref{fig:sens}, right) ranks the link lengths by
$\mathrm{d}\ln W/\mathrm{d}\ln L$: lengthening $c$ and $e$ most reduces wear,
while $h$ and $b$ increase it; $k$ is the feasibility-limiting member
($\pm 2\%$ already breaks assembly at some crank angles) and is excluded from the
wear ranking. As design guidance, $c$ and $e$ are the highest-leverage knobs for
durability, but increases must be balanced against the binding ground-clearance
constraint and the feasibility limit on $k$. A variance-based global Sobol
analysis over a $\pm 10\%$ box about Jansen (Fig.~\ref{fig:sobol}) corroborates
and extends this: link lengths $c$ and $k$ dominate the total-wear variance
(total-effect indices $S_T=\RsobolcST$ and $\RsobolkST$), with $c$ acting almost
entirely through interactions ($S_1=\RsobolcSone$ vs $S_T=\RsobolcST$) and $k$
through a strong direct effect ($S_1=\RsobolkSone$); all other links contribute
marginally.

\subsection{Validation and convergence}
\begin{figure}[htbp]
\centering
\includegraphics[width=0.95\linewidth]{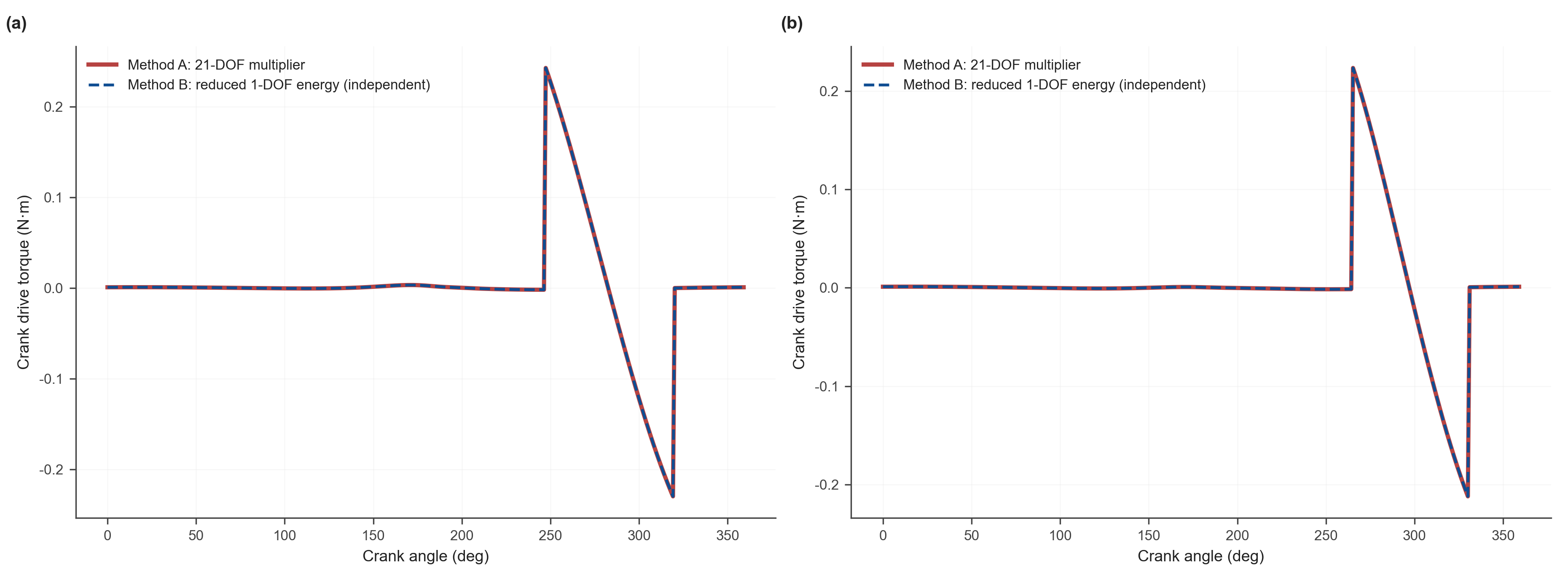}
\caption{Independent cross-verification of the inverse dynamics: crank torque
from the 21-DOF constraint-Lagrange-multiplier solver (method A) and from a
reduced single-DOF energy formulation (method B) coincide to within
$\Rverifytorque$ over the cycle, for both designs: (a) Jansen, (b) optimized.}
\label{fig:verify}
\end{figure}

\begin{figure}[htbp]
\centering
\includegraphics[width=0.6\linewidth]{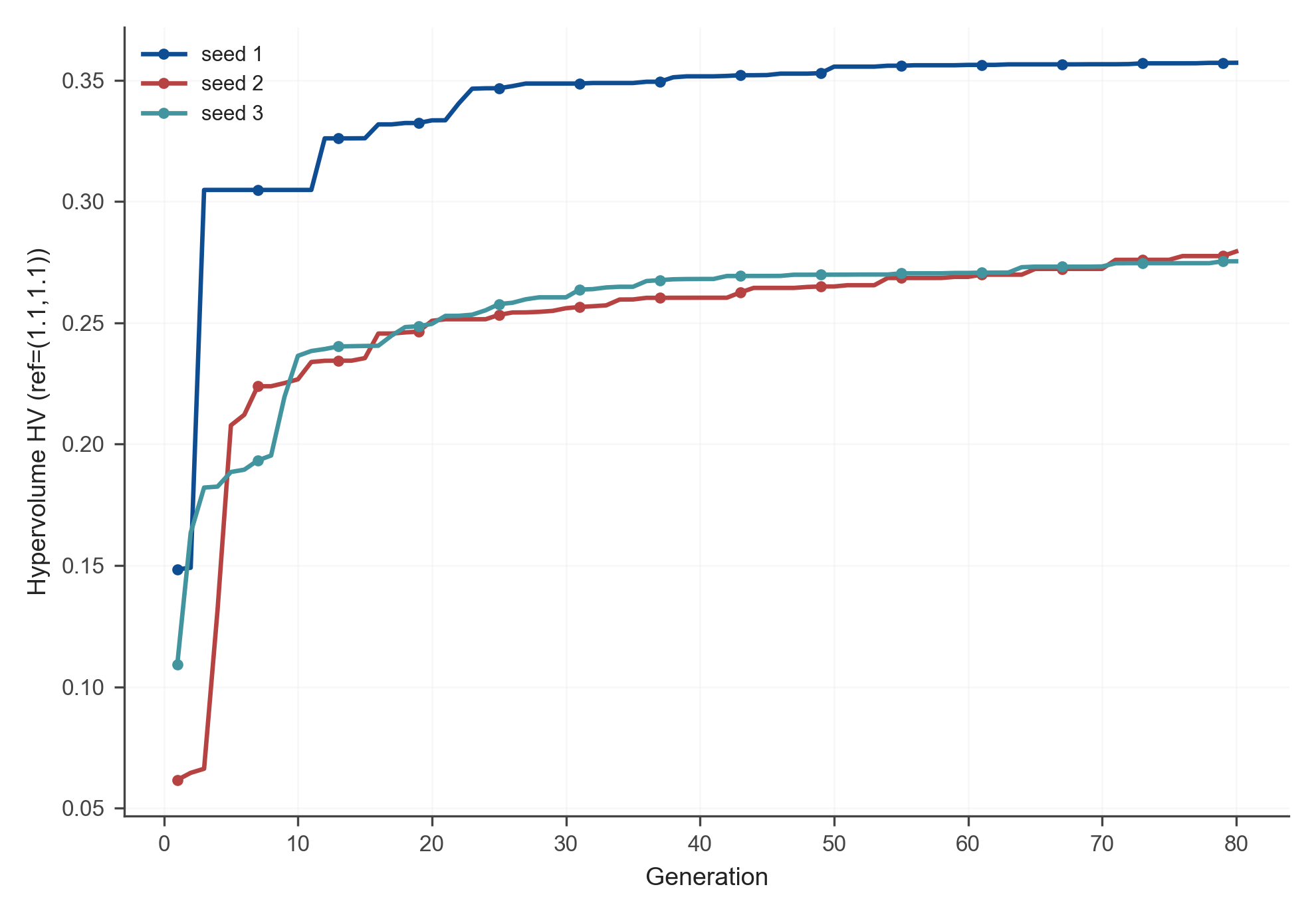}
\caption{NSGA-II convergence: hypervolume vs generation for three independent
seeds (low final-generation variance indicates a converged, repeatable front).}
\label{fig:conv}
\end{figure}

Four checks support credibility in the absence of experiment. (i) The
computed foot path reproduces the canonical Jansen trajectory (flat stance, high
return arc; Fig.~\ref{fig:kin}). (ii) The inverse dynamics is cross-verified by an
\emph{independent} reduced single-DOF formulation: the crank torque from the
21-DOF constraint-Lagrange-multiplier solver and from a Lagrangian in the crank
angle alone (whose only shared inputs are the kinematics and the link masses)
agree to within $\Rverifytorque$ over the entire cycle (Fig.~\ref{fig:verify}),
the instantaneous power balance closes to $\Rverifypower$, and the
cycle-integrated torque vanishes under the conservative load model---an
energy-consistency check. (iii) The optimization is converged and repeatable: across three
independent seeds the hypervolume~\citep{zitzler2003} rises and plateaus (Fig.~\ref{fig:conv}), with
final-generation hypervolumes of $\HVa$, $\HVb$, $\HVc$ (standard deviation
$\HVstd$), and the merged front contains seven non-dominated designs---so the
result is not a single-seed artifact. As a sanity baseline, a uniform random
search over the design box ($\Nrandsamp$ samples, $\Nrandfeas$ feasible) reaches a
hypervolume $\Rhvgain$ below the optimizer's and produces \emph{no} design that
dominates the representative, confirming the front is a genuine optimization
outcome rather than an artifact of dense sampling. (iv) The wear conclusion is independent of
the tribological constants (which cancel in the ratio) and robust to the
mean-force approximation (Section~4). A further external cross-validation of the
joint-reaction forces against a commercial multibody solver (ADAMS/Simscape), for
which the full model specification is provided as supplementary material, is left
to future work.

\begin{figure}[htbp]
\centering
\includegraphics[width=0.95\linewidth]{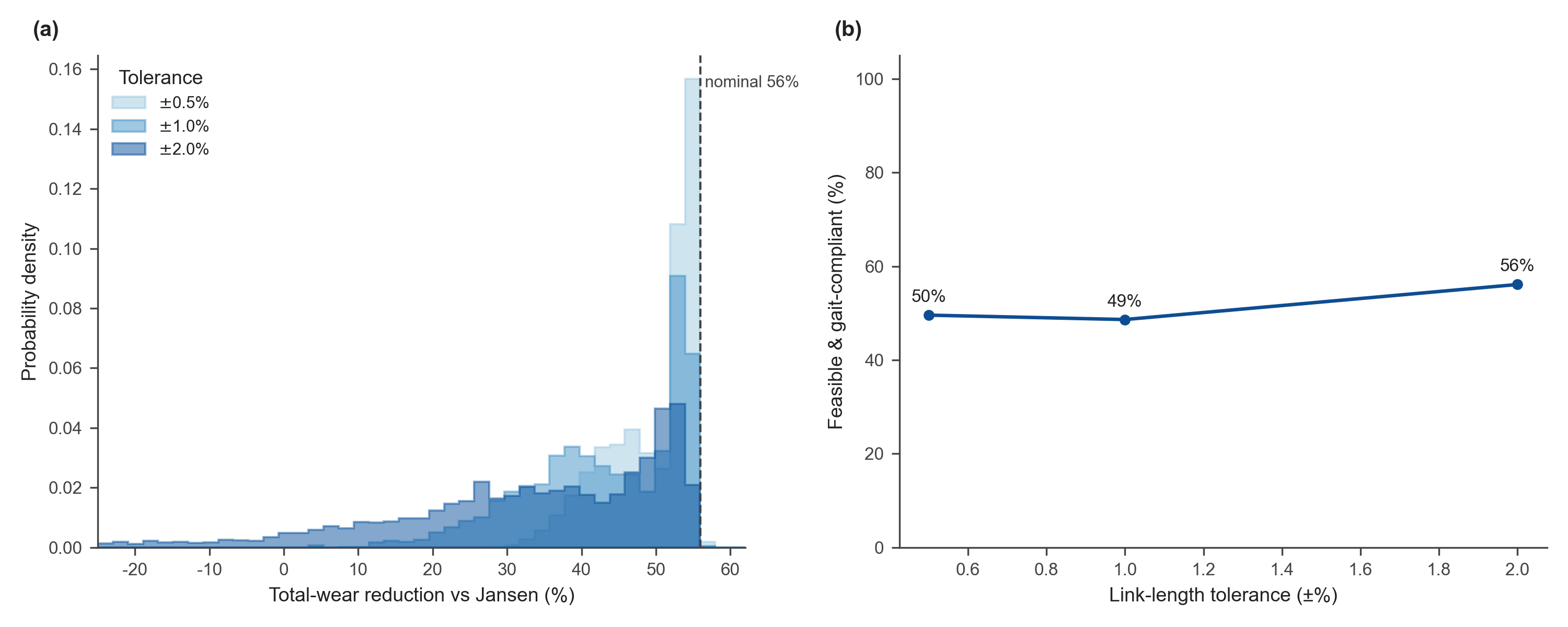}
\caption{Manufacturing-tolerance Monte Carlo: (left) distribution of total-wear
reduction under $\pm0.5/1/2\%$ Gaussian link-length error (feasible samples;
dashed line $=$ nominal $\Rwear$); (right) fraction of toleranced builds that
remain assembly- and gait-feasible.}
\label{fig:tol}
\end{figure}

\textbf{Manufacturing robustness.} A Monte-Carlo tolerance study
(Fig.~\ref{fig:tol}) perturbs the ten optimized link lengths by Gaussian
manufacturing errors. The wear advantage is robust in \emph{magnitude} for builds
that meet the spec: feasible samples retain a mean total-wear reduction of
$\Rtolmean$ at $\pm 1\%$ tolerance (5th percentile $\Rtolpfive$), degrading
gracefully to $\Rtoltwo$ at $\pm 2\%$. However, because the representative design
sits on the active clearance and feasibility boundaries, only $\sim\Rtolfeas$ of
toleranced builds stay within the gait/clearance spec---so the
clearance-governing links require tight tolerancing in practice, a caveat
consistent with the active-constraint discussion above.

\section{Conclusion}

We introduced joint durability into the design of the Jansen linkage by
coupling a parametric kinematic model, a Lagrange-multiplier inverse-dynamic
model and an Archard wear model, and optimizing link lengths for gait quality and
total joint wear with NSGA-II. Under the adopted gait metric the classical ``holy
numbers'' are Pareto-dominated: a $\le\Rlink$ link-length refinement
simultaneously improves gait and cuts total joint wear by $\sim\Rwear$, robustly
across operating conditions. Although demonstrated on the Jansen leg, the
kinematic--dynamic--wear evaluator and the bi-objective formulation are not
mechanism-specific: they apply to any single-DOF leg linkage (e.g.\ Klann or
eight-bar designs), making durability-aware dimensional synthesis a general design
tool. Limitations are the idealized (clearance-free)
joints---so the wear figures are relative rankings, not absolute life
predictions---the mean-force approximation in \eqref{eq:archard}, the dependence
of the Pareto ranking on the adopted composite gait metric (verified stable only
over flatness/ripple weights of $0.3$--$0.7$), and the absence of experiment. The natural extension, and ongoing work, is to replace the
idealized pins with clearance joints and couple the foot--ground impact, so that
wear-induced clearance growth and its feedback on impact dynamics
(via continuous contact-force models~\citep{lankarani1990,marhefka1999,flores2008}) can be
captured.

\appendix
\section{Per-joint wear breakdown}

Table~\ref{tab:perjoint} lists the per-cycle Archard wear at each of the ten
pins for the Jansen and representative optimized designs---the numerical basis for
the per-joint comparison in Fig.~\ref{fig:cmp}. Wear decreases at \emph{all ten}
pins ($-20\%$ to $-85\%$), confirming a broad-based improvement rather than a
single-joint artifact. The most-worn Jansen pin ($G$, $c$--ground) falls by
$82\%$; the optimized peak then shifts to the slip-dominated crank bearing $O$,
whose wear is fixed by the input rotation and cannot be lowered by geometry
(Section~4), so that $O$ sets the residual wear floor.

\begin{table}[htbp]
\centering
\caption{Per-cycle joint wear ($\times 10^{-15}\,\si{\cubic\metre}$ per cycle),
Jansen vs.\ optimized}
\label{tab:perjoint}
\small
\begin{tabular}{lccc}
\toprule
Pin & Jansen & Optimized & Change \\
\midrule
$O$ (crank--ground) & 5.07 & 4.06 & $-20\%$ \\
$J_1$ (crank--$j$) & 5.07 & 4.06 & $-20\%$ \\
$J_1$ ($j$--$k$) & 1.21 & 0.19 & $-85\%$ \\
$J_2$ ($j$--rocker) & 1.58 & 0.93 & $-41\%$ \\
$J_3$ (rocker--$f$) & 1.15 & 0.36 & $-69\%$ \\
$G$ (rocker--ground) & 3.81 & 2.12 & $-44\%$ \\
$G$ ($c$--ground) & 9.90 & 1.79 & $-82\%$ \\
$J_4$ ($k$--$c$) & 1.84 & 0.48 & $-74\%$ \\
$J_4$ ($c$--foot) & 4.42 & 1.19 & $-73\%$ \\
$J_5$ ($f$--foot) & 1.44 & 0.46 & $-68\%$ \\
\midrule
\textbf{Total } & \textbf{35.48} & \textbf{15.64} & $\mathbf{-56\%}$ \\
\bottomrule
\end{tabular}
\end{table}

\FloatBarrier  % ()
\bibliographystyle{unsrtnat}
\bibliography{refs}

\end{document}